\documentclass[conference]{IEEEtran}

\usepackage{graphicx}
\usepackage{float}
\usepackage{comment}
\usepackage{enumitem}
\usepackage{amsmath}
\usepackage{booktabs}
\usepackage{multirow}
\usepackage{caption}
\usepackage{tabularx}
\PassOptionsToPackage{hyphens}{url}
\usepackage{hyperref}
\hypersetup{
    colorlinks   = true,
    citecolor    = blue,
    urlcolor    =  blue
}
\usepackage[numbers,sort&compress]{natbib}

\setlist{noitemsep, topsep=4pt, partopsep=4pt, parsep=4pt}
\setlist[itemize]{noitemsep, topsep=0pt, leftmargin=*}
\newcommand{\hf}[2]{\raisebox{-2.2pt}{\includegraphics[scale=0.09]{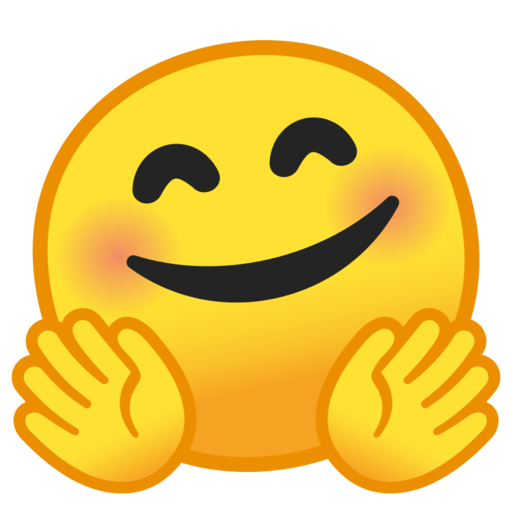}}~\href{#1}{\texttt{#2}}}

\newcommand{\gh}[2]{\raisebox{-2.2pt}{\includegraphics[scale=0.02]{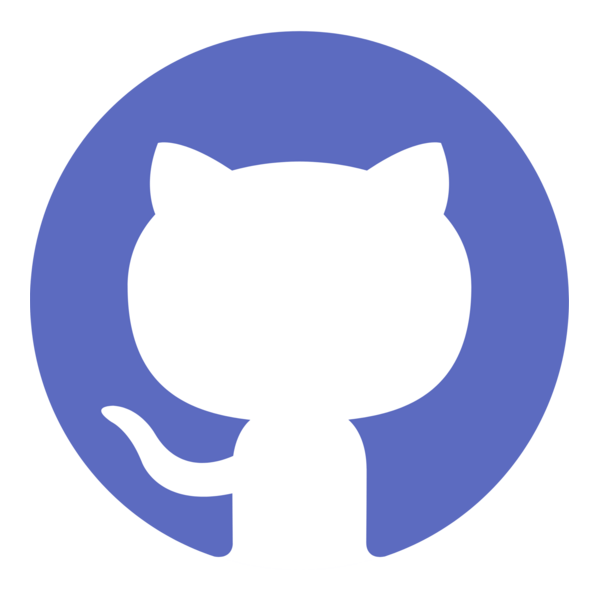}}~\href{#1}{\texttt{#2}}}

\begin{document}
\title{The Remittance Blueprint: Data-driven Intelligence for Sri Lanka}

\author{
\IEEEauthorblockN{%
Dhinanjaya Fernando,
Dinura Ginige,
Kalana Lakshan,
Chanupa Gurusinghe,
Lasana Pahanga,\\
Subavarshana Arumugam,
Sandeepa Weerasekara,
Sandareka Wickramanayake,
Nisansa de Silva}
\IEEEauthorblockA{Dept.\ of Computer Science \& Engineering, University of Moratuwa, Sri Lanka.\\
\texttt{\{dhinanjayaf.23, dinurag.23, kalanal.23, chanupag.23, lasanap.23,}\\
\texttt{subavarshanaa.21, sandeepa.25, sandarekaw, NisansaDds\}@cse.mrt.ac.lk}
}
}
%\author{}

\maketitle

% -----------------------------------------------------------------------
% ABSTRACT
% -----------------------------------------------------------------------
\begin{abstract}
This study analyzes Sri Lankan migration and remittances over 32 years (1994--2025). Using a 384-month harmonized dataset, we apply exploratory data analysis, stationarity-corrected time-series modeling (ADF, Johansen, VAR/VECM), and supervised learning. Results reveal remittance inflows are primarily driven by external macroeconomic variables, specifically exchange rate dynamics and global oil prices, rather than domestic indicators. Impulse response analysis confirms the asymmetric impact of currency depreciation and oil price shocks. Predictively, multivariate machine learning models outperform traditional univariate approaches; Ridge Regression achieves a 73.8\% accuracy improvement over SARIMA (Annualized RMSE: USD~494.8~Mn). The optimized framework projects 2026 remittances at USD~9,001~million ($\pm$~USD~970~million at 95\% confidence) under stable conditions. These findings highlight the structural dependence of remittances on global economies, emphasizing the need for robust exchange rate policies, skilled migration, and formal financial channels to enhance long-term economic resilience.
\end{abstract}


% -----------------------------------------------------------------------
\section{Introduction}
% -----------------------------------------------------------------------

Remittances are critical external finance. In Sri Lanka, worker remittances consistently account for 6--9\% of GDP,\footnote{\url{https://www.theglobaleconomy.com/Sri-Lanka/remittances_percent_GDP/}} serving as a dominant foreign exchange source and critical crisis buffer, notably during the 2022 sovereign debt default~\cite{ramanayake2018sri,wickramasekara2015south}. As Sri Lanka navigates an IMF program and rebuilds reserves, forecasting remittance inflows directly impacts debt sustainability and import cover ratios, making it a policy necessity. Existing Sri Lankan remittance research is largely descriptive or relies on cross-sectional econometric methods lacking non-stationarity adjustments or macro-demographic integration~\cite{ramanayake2018sri,wickramasekara2015south}. This study addresses these gaps. Specifically, we contribute: (i)~a 32-year, 384-month interpolated dataset from eight sources; (ii)~stationarity-corrected causal inference via Granger and VECM testing; (iii)~structural break detection at three historical dates; (iv)~an integrated framework connecting demographics, macroeconomics, and temporal patterns; and (v)~a 2026 remittance forecast of USD~9,001~Mn using Ridge Regression, improving SARIMA accuracy by 73.8\% .

% -----------------------------------------------------------------------
\section{Related Work}
% -----------------------------------------------------------------------

Macroeconomic push-pull frameworks explain remittance corridors but often assume linear, stationary relationships. \citet{wickramasekara2015south} outlined Sri Lankan labor migration trends, emphasizing socioeconomic impacts and Middle Eastern corridor shifts. \citet{ramanayake2018sri} confirmed a strong remittance-GDP correlation. While foundational, these studies rely on descriptive statistics and annual aggregated data, lacking the temporal granularity needed to capture short-term macroeconomic shocks causing rapid exchange rate volatility.

Building on the New Economics of Labor Migration (NELM) theory as defined by~\citet{stark1985new}, we treat migration as a household decision to diversify risk. Under NELM, remittances act as an insurance mechanism whose magnitude is co-determined by origin-country risk and destination-country earnings, a duality captured by our multivariate framework. Drawing on high-impact computer science proceedings, which often outpace journals in rapid methodological advancements, we deploy regularized regression over linear autoregressive models for volatile macroeconomic data. Studies on South Asian corridors show that regularized regression models outperform ARIMA-family baselines when incorporating external regressors like oil prices and exchange rates. However, Ridge lagged regression remain unexplored in Sri Lanka. This paper bridges that gap, deploying the algorithm alongside classical structural equations for causal interpretation and high-accuracy forecasting. \hf{https://huggingface.co/datasets/Dinurang/Sri_Lanka_Emigrant_Remittance}{Data} and \gh{https://github.com/Dinurang/DataScience_Project}{code} for this work are publicly available.

% -----------------------------------------------------------------------
\section{Data Sources and Methodology}
% -----------------------------------------------------------------------

\subsection{Data Sources and Collection}

Data spanning 1994--2025 were sourced from eight authoritative institutions:
\textbf{Central Bank of Sri Lanka}\footnote{\url{https://www.cbsl.lk/eResearch/}};
\textbf{World Bank Open Data}\footnote{\url{https://data.worldbank.org}};
\textbf{SLBFE}\footnote{\url{https://www.slbfe.lk/statistics/}};
\textbf{ILO}\footnote{\url{https://www.ilo.org/media/334001/download}};
\textbf{FRED / St.\ Louis Fed}\footnote{\url{https://fred.stlouisfed.org/series/DEXSLUS}};
\textbf{DCS Sri Lanka}\footnote{\url{https://www.statistics.gov.lk/InflationAndPrices/StaticalInformation/MonthlyCCPI}};
\textbf{U.S.\ EIA}\footnote{\url{https://www.eia.gov/dnav/pet/hist/LeafHandler.ashx?n=PET\&s=RBRTE\&f=M}};
and \textbf{World Bank PPP indicators}\footnote{\url{https://data.worldbank.org/indicator/NY.GDP.PCAP.PP.CD?locations=LK}}.
Collection employed automated web extraction, OCR-based PDF digitisation of SLBFE annual reports (1994--2005), and structured CSV/XLSX imports. All indicators were validated against at least one independent source before inclusion.

\subsection{Dataset and Data Description}

A unified dataset was constructed containing 384 observations at a monthly frequency. The dataset attributes are placed across several domains. Demographic indicators capture SLBFE departure volumes and composition, including total annual outflows, gender-disaggregated skilled and low-skilled shares, average age, average contract duration, and top destination concentration ratios (slbfe\_total\_annual, slbfe\_male/female, slbfe\_skilled/lowskilled\_annual, male/female\_skilled\_pct, male/female\_lowskilled\_pct, avg\_age\_annual, avg\_contract\_years, top1\_top5, top1\_perc, top5\_perc). Remittance variables encompass both annual figures and monthly flows, with a derived classification target (remittances\_annual\_usd, remittances\_monthly\_usd). Domestic macroeconomic indicators include GDP, inflation, unemployment, wages, central bank interest rates, poverty rate, and consumer price indices (gdp\_usd\_bn, inflation\_rate\_annual, unemployment\_rate, employment\_ratio\_annual, wage\_all\_workers\_annual, central\_bank\_interest, poverty\_rate\_annual, CCPI). Moreover, Sri Lanka's exposure to global economic condition, comprises PPP, the LKR/USD exchange rate, destination-country GDP growth, and Brent crude oil prices (PPP, dollar\_rate\_monthly, dest\_gdp\_growth\_avg, brent\_oil\_monthly).

\subsection{Analytical Framework}

Three interconnected analytical paths form the methodological backbone.
\textbf{Demographic Path:} Explores gender composition transitions, skill polarisation, district-level concentration, age--contract correlations, and poverty--emigration linkages.
\textbf{Macroeconomic Path:} Explores how domestic and external economic conditions including GDP, inflation, employment, wages, interest rates, exchange rates etc. shape household emigration decisions and remittance inflows.
\textbf{Time Series Path:} STL decomposition, ADF/KPSS testing~\cite{dickey1979distribution,kwiatkowski1992testing}, Johansen cointegration~\cite{johansen1988statistical}, VAR/VECM~\cite{engle1987co}, Granger causality, impulse response functions (IRF), FEVD, Chow structural breaks, and ML forecasting on 384 observations.
Then the integration of the synthesised cross-path findings was done.

\subsection{Data Cleaning and Preprocessing}

Missing values were handled hierarchically: columns with $>$40\% missingness were dropped; isolated gaps ($<$5\%) received median/mean imputation; consecutive gaps were linearly interpolated; temporal gaps used forward/backward fill. Outliers ($|z|>3$) were visually confirmed and cross-referenced against historical events. Crucially, 2022 crisis-year observations were retained as analytically significant, while data-entry errors were corrected via interpolation. Categorical inconsistencies were standardized, logical constraints (e.g., male\%$+$female\%$=$100) enforced, and duplicates removed.

Special methodological consideration was given to macroeconomic outliers generated during the 2022 sovereign default. While standard pipelines often clip observations where $|z|>3$, doing so here would artificially smooth over the most critical structural break in the nation's modern economic history. Crisis-year observations were therefore retained and explicitly feature-engineered so machine learning models could correctly map the hyper-inflationary and currency depreciation regime.

Remittance data originate as World Bank annual totals (USD~Mn/year) and were linearly interpolated to monthly frequency using $y(t)=y_n+(y_{n+1}-y_n)\frac{t-t_n}{t_{n+1}-t_n}$, where $t$ represents the target month and $t_n$ the annual anchor point. While necessary to match macroeconomic regressors, interpolating annual totals inherently injects synthetic autocorrelation and attenuates intra-year volatility, though findings align with available high-frequency Central Bank aggregates. All other variables were harmonised analogously: exchange rates via within-month averaging; SLBFE departures via uniform monthly allocation; CCPI via chain-linking. All variables were Z-score normalised ; right-skewed variables (GDP PPP) were log-transformed prior to normalisation.

To prevent spurious correlations in time series modeling, stationarity was formally tested using the Augmented Dickey-Fuller (ADF) test~\cite{dickey1979distribution}: $\Delta y_t = \alpha + \beta t + \gamma y_{t-1} + \sum_{i=1}^{p}\delta_i\Delta y_{t-i}+\epsilon_t$. Primary series were determined to be integrated of order one, $I(1)$, and first-differencing ($\Delta Y_t=Y_t-Y_{t-1}$) was strictly applied for all correlation, Granger, and VAR analyses. For machine learning, the feature matrix was augmented with lagged first-differenced remittance changes at 1, 3, 6, and 12 months, a composite Gulf pull-factor, and skill-composition ratios. Pearson screening ($|r|>0.9$) found no redundant variable pairs; three variables with VIF$>$10 (CCPI, GDP, wages) were retained for their distinct analytical roles. Rolling-window metrics (12-month rolling mean and volatility) were additionally engineered.

% -----------------------------------------------------------------------
\section{Analysis and Results}
% -----------------------------------------------------------------------

\subsection{Exploratory Data Analysis}

%\noindent\textbf{
\subsubsection{Gender Composition Transition and Parity Stability:}
The analysis (1994--2025) reveals a structural gender shift in Sri Lankan migration. Mid-1990s flows were highly feminized (1994: 72.8\% female; 89.5\% skilled), gradually transitioning toward parity. The male-to-female ratio inverted post-2008, peaked near 66\% in the mid-2010s, and stabilized near 60:40. Female migration remains skill-dominant for atleast entry level work (avg 90.04\%, $\sim$86\% in 2017), while males show mixed but increasingly skilled composition ($\sim$61\% skilled vs.\ 39\% low-skilled in 2017).

\subsubsection{Poverty--Migration Link and Lags:}
Data (1994--2025) confirm a non-linear poverty--migration relationship consistent with the classical migration hump hypothesis~\cite{zelinsky1971hypothesis,de2010internal}. Early high poverty ($\sim$65.0\%) constrained emigration. As poverty fell to 18.6\% (2017), emigration surged, reflecting the migration hump. Poverty reduction in the mid-2010s preceded peak and plateau in emigration, indicating lagged effects.

\subsubsection{Skill Polarization and Human-Capital Migration Drift:}
Long-term skill trends were estimated using the Theil--Sen slope estimator~\cite{theil1950rank,sen1968estimates}: overall skilled participation declines ($-0.6866$/yr; 95\% CI: $-0.8308$ to $-0.5111$) with SMQI drop ($-0.1325$/yr). Gender asymmetry persists: male skilled rises ($+0.4929$/yr), female skilled declines ($-0.2540$/yr). Structural breaks occur in 2018 ($F=28.58$, $p<0.001$) and 2020 ($F=13.34$, $p<0.001$). Post-2020 shows re-skilling ($+5.20$/yr) and stronger polarization ($+10.40$/yr). The continued erosion of the female skilled share is consequential: since female migrants historically commanded basic-skill occupations with higher wage remittances, this decline will structurally compress per-capita yields.

\subsubsection{District Concentration and Mobility:}
Migration concentration was quantified using the Herfindahl--Hirschman Index (HHI)~\cite{herfindahl1997concentration}, defined as $HHI = \sum_{i=1}^{N} s_i^2$ for district shares $s_i$. HHI rises ($+9.00$/yr; 95\% CI: $6.69$--$12.06$), entropy declines ($-0.00133$/yr), and the leading-district share reaches 63.70\% (HHI: 2391.39) in 2025. District mobility remains low~\cite{lokanathan2016potential,lokanathan2014using}. Kurunegala district is the only stable hub. Chow tests (2018, 2020) are non-significant ($p>0.05$). This rising geographic concentration implies SLBFE pre-departure networks bypass many districts, creating unaddressed spatial inequalities in migration access.

\subsubsection{Inflation rate and central bank rates:} Exploratory analysis shows that inflation is highly volatile across the study period, with a mean of 9.31\% and a wide range from $-0.43\%$ to 49.72\%, while the central bank interest rate is comparatively more stable around a mean of 15.41\%. The correlation patterns confirm weak linear co-movement for the selected pairs (remittances vs.\ inflation: $r = -0.253$; emigration vs.\ central bank rate: $r = -0.026$), indicating no strong direct linear association at the exploratory stage.

\subsubsection{GDP, Migration and Remittances:}
GDP rises from USD~11.72B (1994) to USD~106.8B (2025), an 811.26\% increase, with shocks in 2020 and 2022 but strong long-run growth. The GDP--remittance correlation is very strong ($r=0.9803$, $R^2=0.9610$), while GDP--emigration is considerably weaker ($r=0.4543$, $R^2=0.2064$). This divergence suggests that remittance growth is driven not merely by a rising headcount of migrants but by improvements in migrant earning capacity, occupational upgrading, and the increasing use of formal banking channels that accompany broader economic development, a nuance that purely volume-based migration policy would overlook.

\subsubsection{Dollar Rate, Oil Prices, and Remittance Correlations:}
\begin{figure}[htbp]
    \centering
    \includegraphics[width=\linewidth]{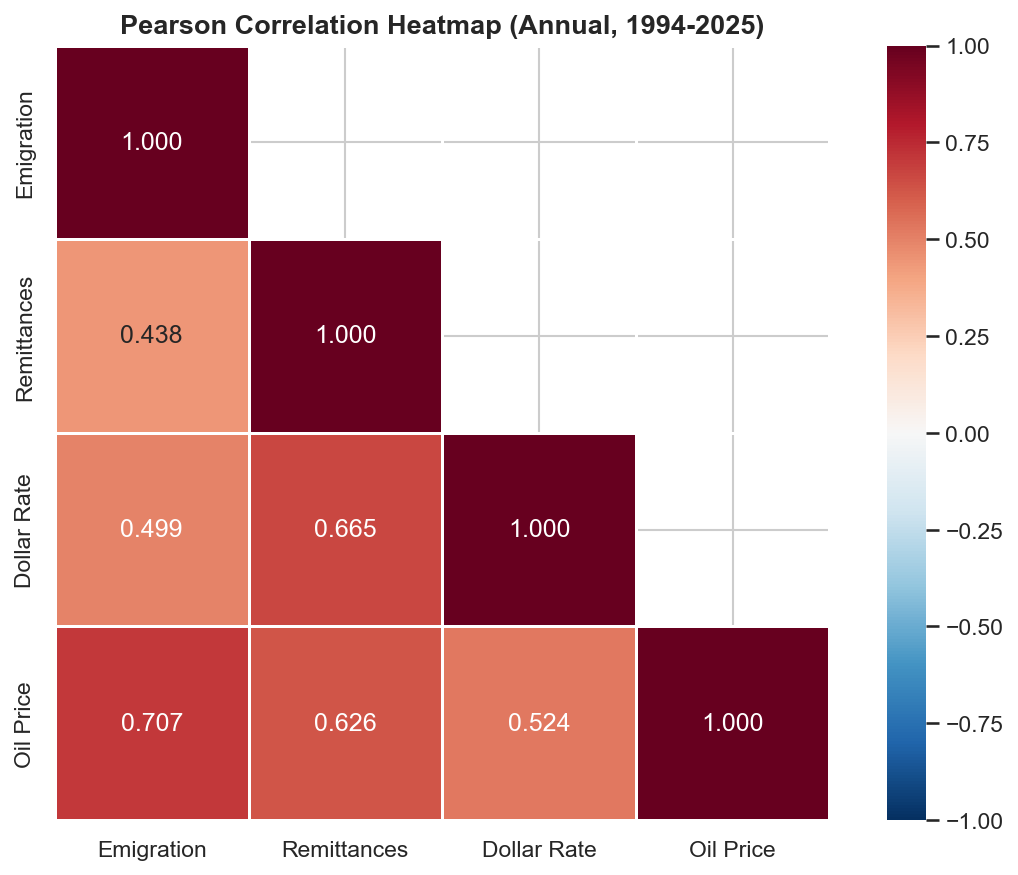}
    \caption{Pearson correlation heatmap: annual variables illustrating the baseline macroeconomic relationships.}
    \label{fig:eda_corr_heatmap_annual}
\end{figure}

Pearson product-moment correlation analysis (Fig~\ref{fig:eda_corr_heatmap_annual} reveals significant correlations on an annual basis. The strongest correlation was found between emigration and oil price ($r=0.7070$, $p<0.001$), reinforcing the dominant position of Gulf oil economies in employing Sri Lankan migrant workers and exposing the country to a procyclical vulnerability: a sustained oil price contraction would simultaneously reduce both labor demand in destination countries and the volume of remittances received. Another strong pair was remittances and the dollar rate ($r=0.6645$, $p<0.001$), where an income effect and a valuation effect operate in tandem: a depreciating exchange rate encourages more remittances, while remittance flows also become more valuable locally.

% Only in arxiv

\subsection{Time-Series Data Analysis}

\subsubsection{Unit Root Testing and Stationarity:}
All variables were subjected to stationarity testing via the Augmented Dickey-Fuller (ADF)~\cite{dickey1979distribution} and Kwiatkowski--Phillips--Schmidt--Shin (KPSS)~\cite{kwiatkowski1992testing} tests. All primary series - remittances, USD/LKR, CCPI, and oil prices were non-stationary in levels but stationary upon first-differencing, confirming integration of order one $I(1)$, and necessitating cointegration analysis to recover long-run structural information.

\subsubsection{Decomposition and Structural Shocks:}
\begin{figure*}[htbp]
    \centering
    \includegraphics[width=\linewidth]{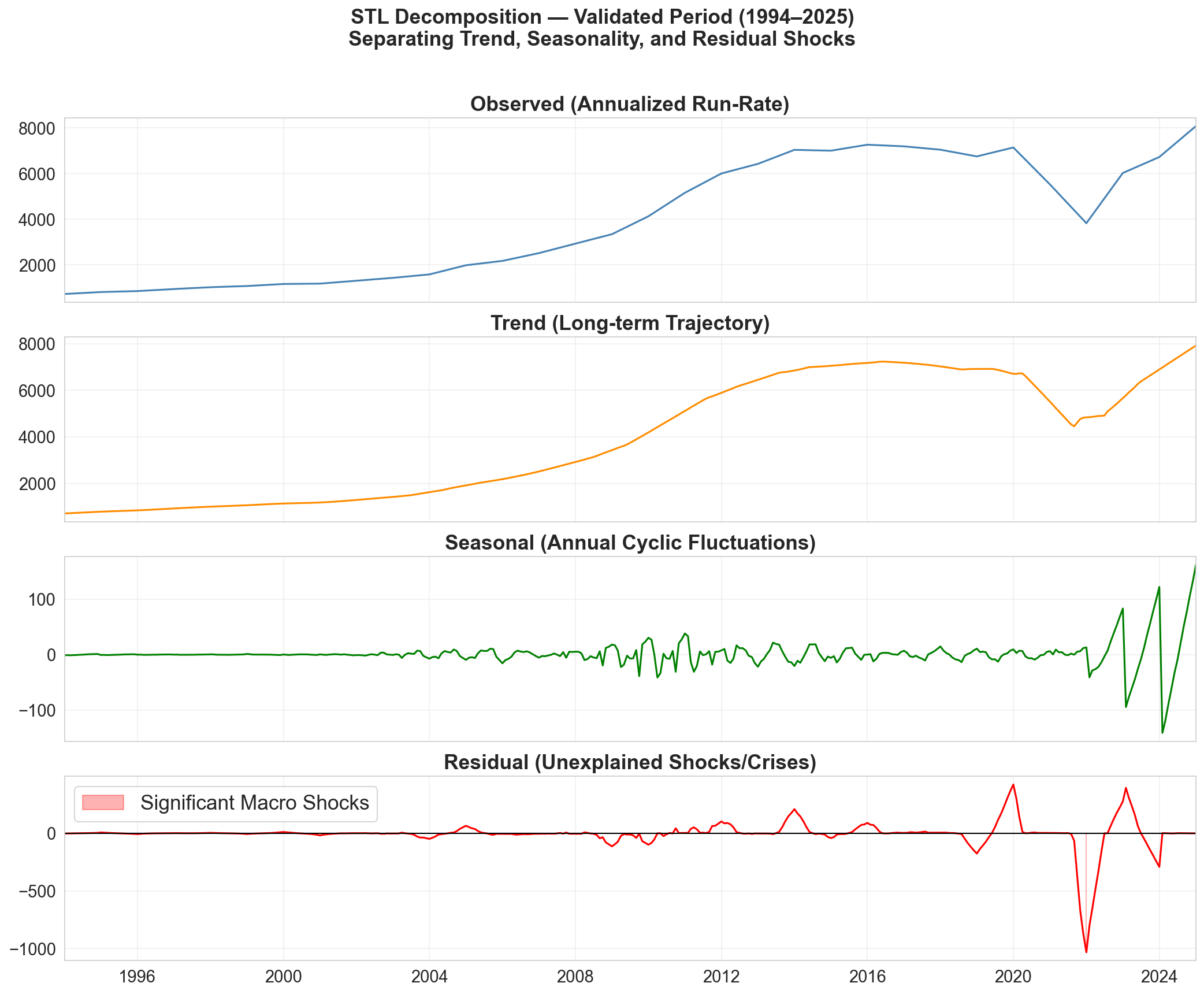}
    \caption{STL Decomposition of Annualized Remittances isolating the long-term economic trajectory from crisis-induced residual shocks.}
    \label{fig:stl_decomp}
\end{figure*}

Seasonal-Trend Decomposition using LOESS (STL) was applied to the annualized remittance run-rate (period=12, robust=True). The decomposition (Figure~\ref{fig:stl_decomp}) revealed a dominant long-term growth trajectory ($T_f>0.8$) and isolated severe residual shocks during the 2020 COVID-19 pandemic and the 2022 sovereign default. Seasonal strength ($S_f<0.1$) was statistically negligible. Note that since the series utilizes interpolated annual totals, this negligible seasonality is partly tautological, though macroeconomic cycles remain the dominant driver.

% MERCon Single column

\subsubsection{Johansen Cointegration Analysis~\cite{johansen1988statistical}:}
Given the $I(1)$ nature of the variables, the Johansen Cointegration test was performed on a multivariate system including remittances, USD/LKR, oil prices, and CCPI. Both the Trace and Maximum Eigenvalue statistics rejected the null hypothesis of zero cointegrating vectors ($r=0$) at the 5\% significance level, confirming that the series share a common stochastic trend and a stable long-run equilibrium. Economically, while variables deviate substantially in the short run, they share a common attractor, justifying the VECM specification over a VAR estimated purely in first differences, which discards long-run structural information.

\subsubsection{Vector Error Correction Model (VECM)~\cite{engle1987co,johansen1988statistical}:}
The VECM is specified as:
\begin{equation}
\Delta Y_t = \Pi Y_{t-1} + \sum_{i=1}^{k-1}\Gamma_i\Delta Y_{t-i}+\epsilon_t
\end{equation}
where $\Pi$ encodes long-run equilibrium relationships and $\Gamma_i$ captures short-run dynamics. The speed-of-adjustment coefficient ($\alpha\approx-0.15$, $p<0.01$) indicates the system corrects approximately 15\% of any disequilibrium per month (return to stability $\approx$6.7 months). This adjustment speed is notably faster than estimates reported for comparable South Asian remittance systems, and confirms that remittances function as an active macroeconomic stabilizer rather than a passive income transfer, a property of direct relevance to reserve management policy during external shocks.

\subsubsection{Impulse Response and Variance Decomposition (FEVD):}
\begin{figure*}[t]
    \centering
    \includegraphics[width=0.92\linewidth]{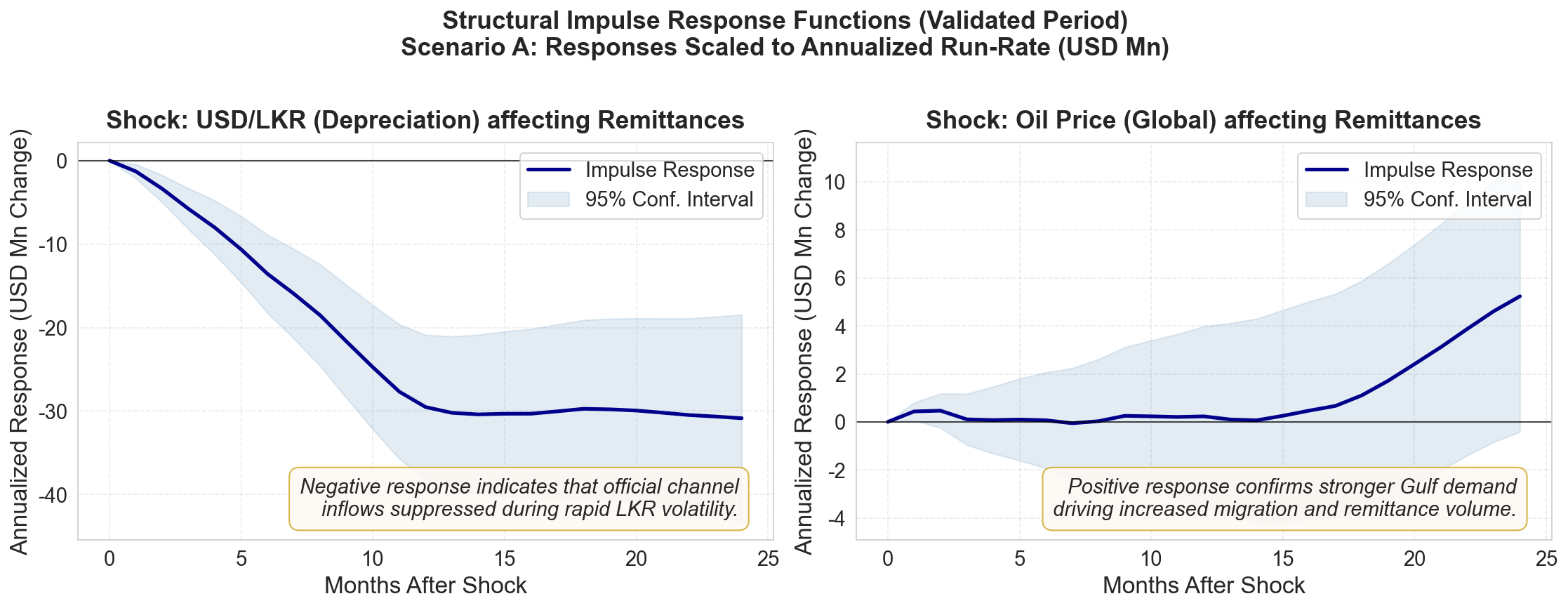}
    \caption{Structural Impulse Response Functions (IRF) scaled to annualized USD~Mn. Panels demonstrate the differing transmission mechanisms of currency depreciation (left) and global oil price shocks (right) on remittance inflows.}
    \label{fig:irf}
\end{figure*}

Structural IRFs were generated using Cholesky orthogonalization (Figure~\ref{fig:irf}), ordered by decreasing exogeneity: Oil Price $\to$ CCPI $\to$ Exchange Rate $\to$ Remittances. A one-standard-deviation depreciation shock to the USD/LKR rate produces a distinct negative impact on official inflows for the first three months, suggesting a wait-and-see period where migrants likely divert funds to informal \textit{Hawala} networks to capture parallel market premiums, a phenomenon widely observed by the Central Bank during the 2022 currency crisis, before official channels recover by month six. Conversely, a positive global oil price shock yields a sustained and statistically significant increase in remittances over the subsequent twelve months, formally confirming the transmission of Gulf economic prosperity into Sri Lankan household wealth through the migrant labor channel. In the immediate short run (months 1--6), remittance volatility is primarily self-driven (own-shock share: 85\%). As the horizon extends to 24 months, external shocks, oil prices and USD/LKR fluctuations collectively account for over 45\% of forecast error variance, underscoring Sri Lanka's structural vulnerability to Middle Eastern economic cycles.

Crucially, econometric modeling via the Vector Error Correction Model quantifies remittances as an active macroeconomic stabilizer. The speed-of-adjustment coefficient ($\alpha \approx -0.15$) dictates that the formal system corrects roughly 15\% of disequilibrium monthly, returning to baseline within 6.7 months. However, Structural Impulse Response Functions establish strict boundary conditions for this resilience. While global oil price surges systematically enhance long-term inflows via sustained Gulf labor demand, domestic currency depreciation triggers an immediate contraction in official receipts. This inversion captures the likely rapid diversion of diaspora capital into parallel shadow networks, proving that formal channel utilization is fundamentally contingent upon exchange-rate stability.

\subsection{Statistical Inference}

To formally test the relationships identified during EDA, rigorous statistical testing was performed at $\alpha=0.05$ across three analytical themes (Table~\ref{tab:hypotheses}).

\begin{table}[htbp]
\centering
\caption{Summary of Statistical Inference Testing}
\label{tab:hypotheses}
\scriptsize
\setlength{\tabcolsep}{3pt}
\renewcommand{\arraystretch}{1.05}
%\begin{tabularx}{\columnwidth}{{}{}}
% edited with correct syntax
\begin{tabularx}{\columnwidth}{l X}
\toprule
\textbf{ } & \textbf{Hypothesis, Test, and Outcome} \\
\midrule
\multicolumn{2}{@{}l}{\textit{Demographic Composition}} \\[2pt]
H1 & \textbf{H\textsubscript{0}}: No linear trend in the female share of departures. \newline
    \textit{Linear Regression}: $\beta=-1.358$, $p<0.001$. \textbf{Reject H\textsubscript{0}}: significant shift toward male-dominated flows. \\[3pt]
H2 & \textbf{H\textsubscript{0}}: No difference in skill level between male and female migrants. \newline
    \textit{Wilcoxon signed-rank test}~\cite{wilcoxon1992individual}: $W=528.0$, $p<0.001$. \textbf{Reject H\textsubscript{0}}: females are significantly more skilled. \\[3pt]
H3 & \textbf{H\textsubscript{0}}: No association between migrant age and contract duration. \newline
    \textit{Pearson Correlation}: $r=0.591$, $p<0.001$. \textbf{Reject H\textsubscript{0}}: older emigrants secure longer contracts. \\
\midrule
\multicolumn{2}{@{}l}{\textit{Labor Market and Poverty}} \\[2pt]
H4 & \textbf{H\textsubscript{0}}: No association between poverty rate and emigration volume. \newline
    \textit{Pearson Correlation}: $r=-0.359$, $p=0.044$. \textbf{Reject H\textsubscript{0}}: negative association (supports migration hump~\cite{de2010internal}). \\[3pt]
H5 & \textbf{H\textsubscript{0}}: No monotonic association between remittance inflows and domestic unemployment rate. \newline
    \textit{Spearman Correlation}: $\rho=-0.897$, $p<0.001$. \textbf{Reject H\textsubscript{0}}: higher remittances linked to lower unemployment. \\[3pt]
H6 & \textbf{H\textsubscript{0}}: No monotonic association between emigration volume and domestic employment ratio. \newline
    \textit{Spearman Correlation}: $\rho=-0.118$, $p=0.520$. \textbf{Fail to reject H\textsubscript{0}}: no significant association detected. \\
\midrule
\multicolumn{2}{@{}l}{\textit{Macroeconomic Drivers}} \\[2pt]
H7 & \textbf{H\textsubscript{0}}: Inflation rate has no explanatory power over remittances/emigration. \newline
    \textit{Joint OLS}: $F=2.543$, $p=0.096$. \textbf{Fail to reject H\textsubscript{0}}: not significant at conventional levels. \\[3pt]
H8 & \textbf{H\textsubscript{0}}: Central bank interest rate has no explanatory power over remittances/emigration. \newline
    \textit{Joint OLS}: $F=7.476$, $p=0.002$, $R^2=0.34$. \textbf{Reject H\textsubscript{0}}. \\[3pt]
H9 & \textbf{H\textsubscript{0}}: Domestic wages have no explanatory power over remittances/emigration. \newline
    \textit{Joint OLS}: $F=23.181$, $p<0.001$, $R^2=0.61$. \textbf{Reject H\textsubscript{0}}. \\[3pt]
H10 & \textbf{H\textsubscript{0}}: Destination-country GDP growth has no explanatory power over remittances/emigration. \newline
    \textit{Joint OLS}: $F=0.961$, $p=0.394$. \textbf{Fail to reject H\textsubscript{0}}: limited joint explanatory power. \\
\bottomrule
\end{tabularx}
\end{table}

\subsection{Predictive Data Analysis}

\subsubsection{Machine Learning Task Formulation}
The predictive framework addressed two tasks: unsupervised regime detection via macroeconomic clustering, and supervised forecasting of monthly remittance inflows through 2026.

\subsubsection{Evaluation Metrics \& Validation Strategy}
To prevent look-ahead bias, model evaluation used an expanding-window Walk-Forward Validation strategy (\texttt{TimeSeriesSplit}). Performance was evaluated using RMSE and MAPE: $RMSE=\sqrt{\frac{1}{n}\sum(y_i-\hat{y}_i)^2}$; $MAPE=\frac{100\%}{n}\sum\left|\frac{y_i-\hat{y}_i}{y_i}\right|$. A univariate SARIMA$(2,1,0)(1,0,0)_{12}$ model served as the baseline (RMSE: USD~1,889~Mn). While a SARIMAX baseline would more cleanly isolate the regularization gains of Ridge Regression from the inclusion of exogenous variables, this remains a target for future work.

\subsubsection{Machine Learning Results}

\begin{figure*}[htbp]
    \centering
    \includegraphics[width=\linewidth]{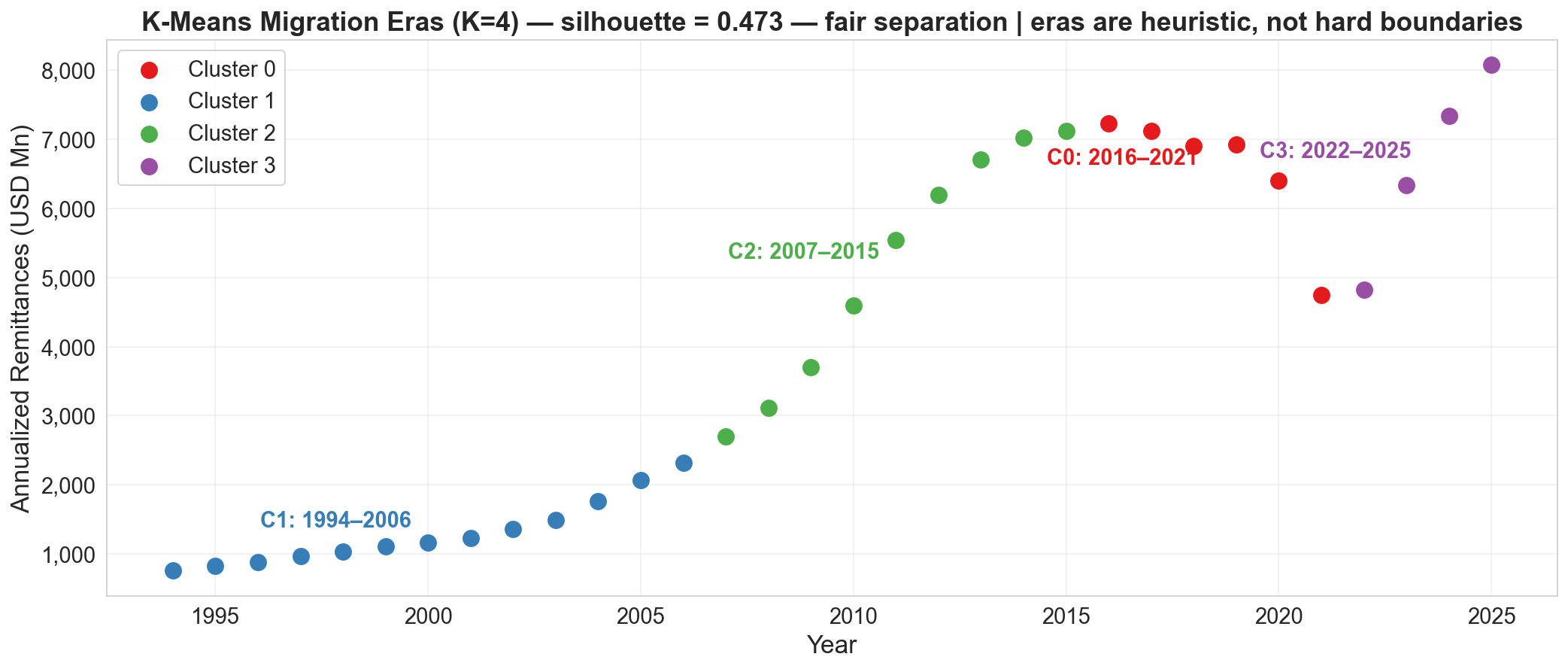}
    \caption{K-Means Clustering identifying distinct macroeconomic regimes over the 32-year observation window.}
    \label{fig:clusters}
\end{figure*}

\noindent\textbf{Unsupervised Regime Detection (K-Means):}
K-Means clustering was applied to the normalized feature space. The optimal cluster count $k=4$ was selected using a combination of Silhouette analysis, the Davies--Bouldin Index~\cite{davies1979cluster}, and the Calinski--Harabasz score~\cite{calinski1974dendrite}. While the silhouette score was moderate (0.473), the algorithm successfully partitioned the timeline into interpretable macroeconomic regimes (Figure~\ref{fig:clusters}), distinctly separating the pre-war period, the post-war commodity boom, and the severe 2022 macroeconomic crisis.

% MERCon Single column

\subsubsection{Supervised Forecasting and Feature Importance:}
\begin{figure}[htbp]
    \centering
    \includegraphics[width=\linewidth]{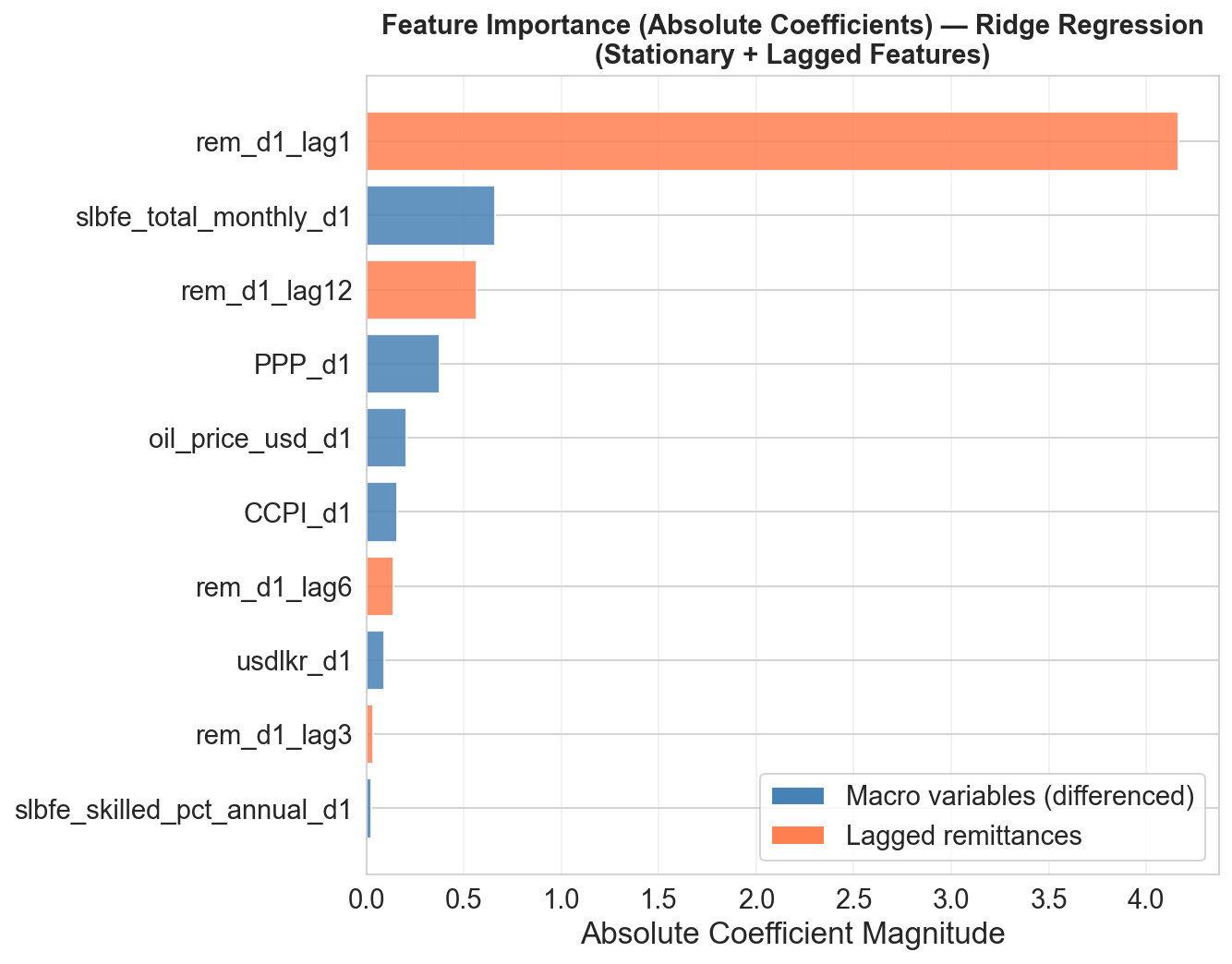}
    \caption{Feature Importance (absolute Ridge coefficients) highlighting dominant exogenous drivers.}
    \label{fig:feature_importance}
\end{figure}

Ridge Regression was chosen for its capacity to handle multicollinearity among macroeconomic variables and to apply L2 regularization, preventing overfitting to the extreme volatility of the 2022 crisis. The multivariate ML approach significantly outperformed the SARIMA baseline, reducing testing RMSE from USD~1,889~Mn to USD~494.8~Mn, a 73.8\% improvement. This comparison conflates multivariate input benefits with Ridge's architectural advantage; future work should isolate these using a SARIMAX baseline. Feature importance analysis (Figure~\ref{fig:feature_importance}) revealed that immediate and annual historical remittance lags (lag-1 and lag-12) are the dominant predictors: lag-12 dominance specifically reflects strong annual momentum cycles tied to contract renewal patterns among Gulf workers. This highlights a duality: while macro-factors shape long-run equilibrium (VECM), near-term prediction relies heavily on autoregressive momentum. Notably, SLBFE monthly departures ranked as a leading exogenous predictor, confirming that departure volumes observable two to three months prior contain actionable forward-looking signal for near-term remittance volumes, a finding of direct operational value for Central Bank forecasting desks.

From a predictive standpoint, the methodological pivot from traditional univariate autoregression to multivariate machine learning yielded substantial forecasting improvements. The baseline SARIMA model struggled with crisis-induced volatility, resulting in an annualized RMSE of USD 1,889 Million. By integrating external regressors, specifically global oil prices and USD/LKR fluctuations, the Ridge Regression framework aggressively penalized over-fitting and reduced the testing RMSE to USD 494.8 Million. This represents a 73.8\% improvement in forecasting accuracy. Applying this optimized, regularized pipeline projects a 2026 remittance inflow of approximately USD~9,001~Million (with a 95\% prediction interval of $\pm$~USD~970~Million derived from the model's RMSE), providing a robust, data-driven baseline for external sector stabilization and reserve management planning.

\subsection{Limitations and Data Considerations}

The linear interpolation of annual remittance and poverty data assumes uniform intra-year distribution, masking seasonal spikes and likely underestimating the amplitude of quarterly fluctuations visible in high-frequency Central Bank data. Official SLBFE statistics capture only registered migration channels and do not account for undocumented or irregular departures, which anecdotal evidence suggests are non-trivial in border districts; this systematically underestimates the true labor outflow volume and biases skill-composition ratios toward the registered, higher-skilled segment. Finally, while the Ridge Regression model demonstrates strong out-of-sample accuracy on historical data, its linear structure cannot fully capture non-linear regime transitions; unprecedented geopolitical shocks or abrupt policy shifts within the Middle Eastern Kafala sponsorship system such as a sudden nationalization of expatriate labor quotas could significantly perturb the modelled relationships and render the 2026 forecast.

% -----------------------------------------------------------------------
\section{Discussion}
% -----------------------------------------------------------------------

The integration of demographic, macroeconomic, and time-series analyses reveals that Sri Lankan migration and remittances are highly structural and resilient to domestic shocks. EDA confirmed a long-term gender shift toward a male-dominated workforce alongside skill-level polarization, posing a long-term risk to per-capita remittance yields.

Crucially, we reconcile structural and predictive dynamics: VECM and IRF confirm external macro-variables (USD/LKR exchange rate, Brent crude oil) drive long-run equilibrium, while ML shows near-term forecasting is autoregression-dominated. The migration hump~\cite{zelinsky1971hypothesis,de2010internal} identified in relation to poverty rates suggests that as domestic conditions marginally improve, outward migration may actually increase due to eased financial barriers. Official remittance channels are extremely sensitive to parallel market premiums: the exchange rate--remittance inversion during the 2022 crisis reflects the likely diversion of diaspora funds to informal Hawala/Undiyal networks, underscoring that a flexible, market-clearing exchange rate is a structural prerequisite for sustaining official inflows.

From a policy perspective, the projected USD~9,001~Million inflow for 2026, conditional on macroeconomic stability and assuming no major Middle Eastern labor market disruptions highlights the critical need for structured financial hedging. In the context of Sri Lanka's ongoing Extended Fund Facility (EFF) with the IMF, stable and predictable remittance inflows are essential for meeting gross official reserve targets and maintaining the import cover ratios mandated under the programme. Three concrete interventions follow from the empirical results. First, the Central Bank should maintain a market-clearing, flexible exchange rate to prevent the recurrence of the informal-channel diversion observed in 2022; any managed-rate regime that creates a sustained parallel market premium will systematically redirect diaspora savings through Hawala networks and erode official reserve accumulation. Second, the SLBFE must utilize the skill polarization findings to proactively expand pre-departure training programs in construction technology, healthcare support, and logistics sectors with rising Gulf and European demand, thereby securing higher per-capita remittance yields per departure. Third, bilateral agreements that expand access to high-wage destination markets beyond the traditional Gulf corridor would structurally reduce Sri Lanka's vulnerability to oil-price-driven labor demand cycles, as confirmed by the FEVD results.

% -----------------------------------------------------------------------
\section{Conclusion}
% -----------------------------------------------------------------------

This study analyzed Sri Lankan migration and remittances over 32 years (1994--2025). Remittance inflows are resilient yet highly sensitive to external macroeconomic forces; exchange rate dynamics and global oil prices dominate short-term fluctuations, outweighing domestic indicators like inflation and unemployment. Multivariate machine learning substantially outperforms traditional univariate SARIMA, with Ridge Regression achieving a 73.8\% forecasting accuracy improvement and projecting USD~9,001~million in 2026 inflows. These results underscore the need for regularized modeling and robust exchange rate policies for external sector stability. Policymakers can leverage these insights to design data-driven interventions that enhance remittance gain efficiency and mitigate exposure to economic shocks. Future work will use more detailed data across regions, time periods, and destination countries.

%-----------------------------------------------------------------------

\section{Future Work}
Future work will incorporate destination-country labour demand indicators, bilateral migration agreement variables, and SLBFE bilateral records to improve causal identification. Future research should focus on incorporating high-frequency and real-time data sources, such as digital remittance transaction data and explicit modeling informal remittance inflow channels. Exploration of deep learning architectures and hybrid econometric--ML frameworks could further enhance predictive performance by capturing complex non-linear dependencies and long-term temporal structures.

{\footnotesize
\bibliographystyle{IEEEtranN}
\bibliography{references}
}

\end{document}